\documentclass[runningheads]{llncs}
\usepackage{times}
\usepackage{latexsym}
\usepackage{framed} 
\usepackage{url}
\usepackage{graphicx}
\usepackage{microtype}
\usepackage{multirow}
\usepackage{multicol}
\usepackage{amsmath}
\usepackage{booktabs}
\usepackage{color}
\usepackage{caption}
\usepackage{enumerate}
\usepackage{enumitem}
\usepackage{subfigure}
\usepackage{arydshln}
\usepackage{setspace}
\usepackage{wrapfig}
\begin{document}
\title{Semi-supervised Visual Feature Integration for Pre-trained Language Models}
%
%
\author{Lisai Zhang \and
Qingcai Chen \and
Dongfang Li \and
Buzhou Tang}

\institute{Harbin Institute of Technology, Shenzhen \\
\email{LisaiZhang@foxmail.com}\\
\email{qingcai.chen@hit.edu.cn}\\
\email{ngly@gmail.com}\\
\email{tangbuzhou@hit.edu.cn}}
\maketitle              
\begin{abstract}
Integrating visual features has been proved useful for natural language understanding tasks. Nevertheless, in most existing multimodal language models, the alignment of visual and textual data is expensive. In this paper, we propose a novel semi-supervised visual integration framework for pre-trained language models. In the framework, the visual features are obtained through a visualization and fusion mechanism. The uniqueness includes: 1) the integration is conducted via a semi-supervised approach, which does not require aligned images for every sentences 2) the visual features are integrated as an external component and can be directly used by pre-trained language models. To verify the efficacy of the proposed framework, we conduct the experiments on both natural language inference and reading comprehension tasks. The results demonstrate that our mechanism brings improvement to two strong baseline models. Considering that our framework only requires an image database, and no not requires further alignments, it provides an efficient and feasible way for multimodal language learning.

\end{abstract}
\section{Introduction}
\begin{figure}
	\vspace{-15pt}
	\renewcommand\tabcolsep{3.0pt}
	\begin{tabular}{p{0.68\linewidth}r}
		\textit{\textbf{Passage:} Today my family is holding a feast in the dinning room to celebrate Christmas, we are going to eat a big cake after singing the Christmas song. Grandpa is a cook, he would make a cake every Christmas.} & \multirow{3}{\linewidth}{\includegraphics[width=0.3\linewidth]{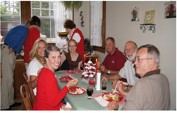}}\\
		\textit{\textbf{Question:} 
			Where is the cake probably placed?} &\\
		\textit{\textbf{Choice:} 
			A. Table B. Oven C. Butter.} & \\
	\end{tabular}
	\vspace{-5pt}
	\caption{A reading material and visualized image.}
	\vspace{-15pt}
	\label{fig:img}
\end{figure}

Human's understanding of natural language and cognition of the objective world is based on the combination of language and various sensory organs. To understand texts, we need to learn the basic knowledge about the real world through observation. For example, Given the following reading comprehension task in Fig.~\ref{fig:img}, 
human may recall a visual scene like the right picture while reading, and easily select the correct answer "A". 
However, most reading comprehension tasks do not provide aligned images, and it is hard for a machine to select without the common sense that a cake is usually placed on a table, which may be described in some pictures. If we can provide the image in Fig. 1 just like human’s visualization, and integrate the information about a cake and a table to NLU models, it is likely to be helpful.

\begin{figure}
	\vspace{-5pt}
	\centering
	\subfigure[\small{A girl standing alone on the river \textbf{bank}}]{
		\includegraphics[width=0.3\linewidth, height=2.3cm]{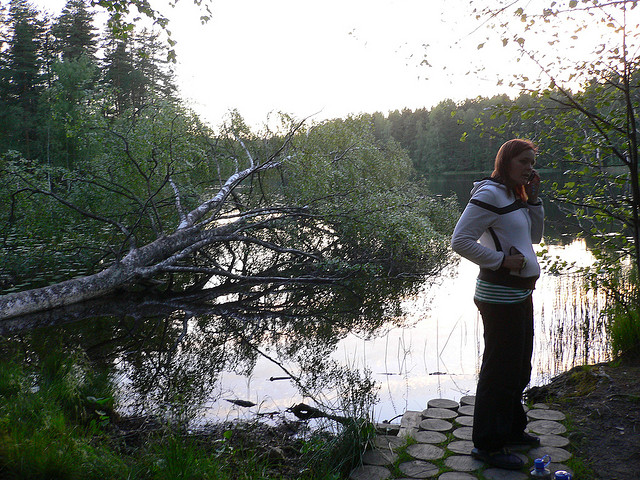}
	}
	\subfigure[\small{\textbf{Bank} office building in the center of street.}]{
		\includegraphics[width=0.3\linewidth, height=2.3cm]{img/example/bank2.png}
	}
	\subfigure[\small{A \textbf{bank} of clouds build up along the western horizon.}]{
		\includegraphics[width=0.3\linewidth, height=2.3cm]{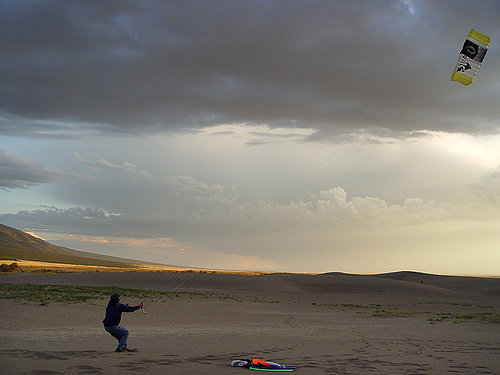}
	}
	\caption{Visualization of sentences with ambiguous word.}
	\label{fig:example}
	\vspace{-15pt}
\end{figure}
Natural language processing researchers have attempted to help machines to understand text using visual information for a long time. Works on word sense and word embedding ~\cite{kiela2014learning,modal_AI:2014} have shown that a better word representation can be learned from multimodal information, but these works are limited by the scale of word-image alignments and only provides multimodal representation for a specific set of words. To achieve large scale word-visual feature fusion, picture-book \cite{hinton:2018} use embedding of 2.2 million words learned from images searched by google resulted in promising results on several tasks when concatenated with word embedding learned from the text. Despite the success of integrating visual features into word embedding, there is a shortage of word-level multimodal representation learning: an image usually provides one specific illustration for the word sense, but a word may have ambiguous meaning. Fig. \ref{fig:example} shows an example of such an issue: by providing a specific image for the word \textit{bank} in a sentence, the ambiguity of word sense disappears, which may be helpful for word sense disambiguation, a big issue for the general-purpose word embedding learning.  

Recently, several sentence-level multimodal semantic learning methods have also been proposed. In the early stage, the main purpose of these models are for generating image captions ~\cite{socher_sentence:2014,sentence_level_multimodal:2015}  and image retrieval ~\cite{hierarchical:2017}. These works are still limited by the scale of aligned data. A recent research ~\cite{2018:NAACL_sentence_visual} transfer image retrieval model to NLP tasks, and get better performance than text-only models on some NLP tasks including entailment and classification. In these works, the language models are learned from aligned text-visual data, so it is hard to apply these models to the pre-trained language models~\cite{devlin:bert,gpt:2018,elmo:2018} which need to be trained on large text corpus.

Unique to existing works, this paper aims to bring visual information through a visualization like mechanism. The approach is semi-supervised and can be integrated to pre-trained NLU models. The main contributions of this paper are outlined as follow:  

\begin{itemize}[leftmargin=*]
	\setlength{\itemsep}{0pt}
	\setlength{\parsep}{0pt}
	\setlength{\parskip}{0pt}
	\item proposes a sentence visualization  and fusion mechanism that provides visual features for every sentence dynamically. Since the mechanism decouples aligned multimodal data and multimodal NLU models, it provides a feasible way for multimodal sentence representation.   
	\item proposes a natural language inference model by integrating visualization mechanism to BERT \cite{devlin:bert}. The integrated model brings performance improvement on SNLI dataset. 
	\item proposes a reading comprehension model by integrating the visualization mechanism to the pre-trained TriAN~\cite{trian:2018} model. On the SemEval 2018 Task 11 the mechanism brings absolute performance gains of 1.28\% and 0.57\% for single and ensemble models respectively.  
\end{itemize}

\section{Related work}
\textbf{Multimodal Representation Learning}
Early language comprehension methods ~\cite{multimodal:2012,autoencoder:2014,kiela2014learning,modal_AI:2014} inject visual information into language model by learning visual-semantic representation. Generally, these researches feed image features to word embedding learning algorithms. More recently, Zablocki et al.~\cite{modal_aaai:2018} proposed to unify text and vision by simultaneously leveraging textual and visual context to learn multimodal word embedding. The work closely to ours is that of Kiros et al.~\cite{hinton:2018} who use google image search as a tool to get images for texts and evaluate the method on several natural language processing tasks. However, they still use images to learn visual-word representation.
Apart from words, sentence level visual-semantic representation learning ~\cite{AK:Devise,AK_multimodal:2015,socher_sentence:2014} deal with the problem as sentence-image ranking learning, they use convolutional neural networks to encode image and recurrent neural networks to encode sentences. To achieve better image-text alignment, hierarchical models ~\cite{AK:reguion_nips,hierarchical:2017} use regions of image objects and syntax of sentence to learn multimodal representation. Faghri et al.~\cite{vse++:2017} achieved best performance on text-image retrieval task by learning joint embedding of text and image with negative sampling.
Multimodal representation learning inject visual information into language models by learning a implicit representation injection, they mainly focus on word similarity and visual-text retrieval tasks. Our proposed imagination mechanism explicitly provide the visual scene of sentences to language model, and give information for language comprehension by visual-text feature extraction. 

\textbf{Visual-Semantic Embedding}
Existing research on sentence-level visual-semantic embedding learning aim to map both images and their captions into a common space ~\cite{AK:Devise,AK_multimodal:2015}. Niu et al. ~\cite{hierarchical:2017} use HM-LSTM to automatically learn representations for phrases and image regions. Faghri et al.~\cite{vse++:2017} achieved outstanding performance on text-image retrieval task by learning joint embedding of caption and image with hard negative sampling.

\textbf{Visual Question Answering}
The visual question answering research  ~\cite{vqa:2014} is also related to our work because they have similar usage of image with us: use deep models to reason about linguistic and visual inputs jointly. This task has seen improvement since recent years by different approaches, including designs of attention mechanism ~\cite{vqa:2016,vqa_att:2016,vqa:2018}, learning different neural network structures ~\cite{vqa_naacl:2016,vqa_icml:2016} and implementation of multimodal representation \cite{vqa_survey:2017}. Recently, Teney et al.~\cite{vqa:2018} proposed bottom-up and top-down attention mechanism,  and achieved impressive improvement. 

\section{Approach}
\begin{figure}
	\centering
	\includegraphics[width=0.9\linewidth]{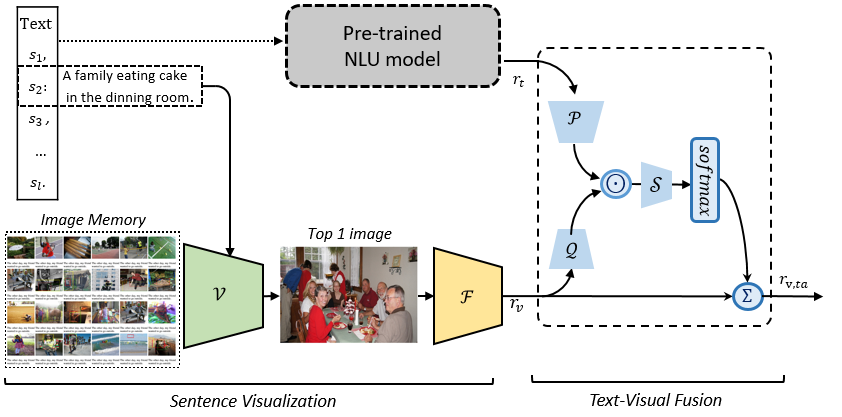}
	\caption{Architecture of the sentence visualization and fusion framework. $\mathcal{V}$ is text to image retrieval model, $\mathcal{F}$ is visual feature extractor. The pre-trained NLU model could be any  language models.}
	\vspace{-10pt}
	\label{fig:imagination}
\end{figure}
This sentence visualization and fusion framework is formulated as two parts: 1) sentence visualization, and 2) the text-visual fusion. The architecture is illustrated in Fig \ref{fig:imagination}.  For a given task that may input with passages or text, the fusing mechanism will retrieval and generate visual representations sentence by sentence, then fusing text and visual features with the attention of the whole text representation.

We define the text of input as a set of sentences, i.e., $\mathcal{T}=\{s_1, s_2, ..., s_l\}$, where $s_i$ is the $i^{th}$ sentence within the text $\mathcal{T}$, and $l$ is the length of $\mathcal{T}$ counted by sentences.  

\subsection{Sentence Visualization}
\label{sec:Visualization}
The visualization provides visual information for a given sentences.
The input text is feed into text encoder and converted to $t \in \mathbf{R}^{J}$ where $J$ is the joint embedding size. Image encoder computes all images in image memory as image embedding $V \in \mathbf{R}^{J \times M}$ where $M$ is the size of image memory base. Then the module return the top 1 result of matched images to Visual Feature Extraction Module as Eq.~\ref{eq:sentence_visualization}
\begin{equation}
\begin{aligned}
\mathcal{V}(t) & = \operatorname*{arg\,max}_{ \theta \in [0, m]} V_{\theta} \cdot t. \\
\end{aligned}
\label{eq:sentence_visualization}
\end{equation}
where $m$ is the imagebase size.
In implement, we use 152-layer-ResNet~\cite{resnet} as image encoder and bidirectional GRU as text encoder.

Two extractors for different levels of visual features are applied. A pre-trained Faster R-CNN network is used to extract object level features, which select top K possible objects from given image and return their object features $I \in \mathbf{R}^{K \times \hat{V}}$ where $K$ is number of objects, $\hat{V}$ is dimension of visual features. A ResNet-152 network pre-trained on ImageNet is used to extract the global features. 

\subsection{Text-Visual Fusion}
\label{sec:Fusion}
After obtaining visual features $r_v$, the framework combines them with the text representation $r_t$ from the pre-trained language model. An attention net $\mathcal{S}$ is is proposed to highlight the key information in the visual features. Each visual feature vector computes an attention weight with text representation as Eq.~\ref{equ:att}:
\begin{equation}
\alpha = softmax(\mathcal{S}(\mathcal{Q}(r_v) \odot \mathcal{P}(r_t)))
\label{equ:att}
\end{equation}
where $\alpha$ is attention weight on $r_v$, $\mathcal{Q}$ and $\mathcal{P}$ are one hidden layer DNN with weighted norm and ReLU activations, $\mathcal{S}$ is a linear transform, $\odot$ is the Hadamard (element-wise) product.
Finally sum $r_v$ is weighted and one visualization vector $r_{v,ta}$ is output as Eq.~\ref{equ:vta}.
\begin{equation}
r_{v,ta} = \sum_{i=1}^{N}{r_{i,v} \times \alpha_i}
\label{equ:vta}
\end{equation}
where $N$ is the 
The $r_{v,ta}$ is used as the text-visual fusion representation and could be fed into corresponding NLU models.

To demonstrate the efficacy of visualization mechanism for different grains of text, we propose two NLU tasks: one is the natural language inference that takes two sentences as input, another is the reading comprehension task that takes a passage of text as given material, a question and several choices as input.  Following two parts introduce integrating the visualization  models for these two tasks respectively.

\subsection{Semi-supervised Integration}
The visualization mechanism is integrated into pre-trained NLU models through three-steps:
Firstly, train the visualization network in Sec~\ref{sec:Visualization} on aligned text-image data.
Secondly, combine the visualization network with the pre-trained NLU model through the text-visual fusion network in Sec~\ref{sec:Fusion}, set gradient of the pre-trained NLU model to zero and fine-tune the visualization mechanism part only.
Thirdly, fine-tune the whole integrated model end-to-end.

\subsubsection{Integration to Language Inference}
\begin{figure}
	\centering
	\subfigure[Integrated NLI model.]{
		\includegraphics[width=0.48\linewidth]{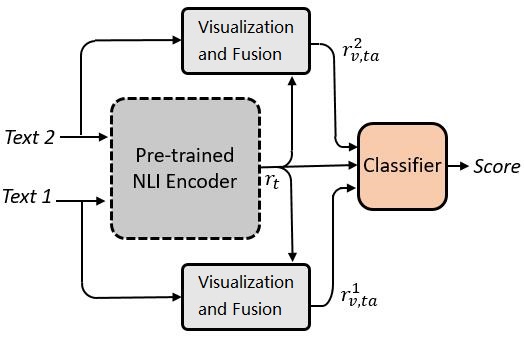}
		\label{fig:inference}
	}
	\subfigure[Integrated reading comprehension model.]{
		\includegraphics[width=0.48\linewidth]{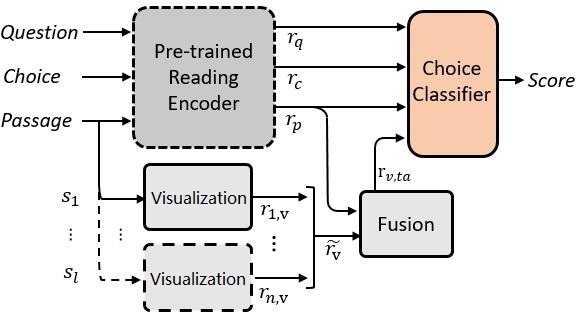}
		\label{fig:MRC}
	}
	\caption{Integration to pre-trained natural language inference model and reading comprehension model.}
	
	\vspace{-10pt}
\end{figure}

In Fig.~\ref{fig:inference}, we give an overall structure of integrating visualization mechanism to natural language inference model such as BERT~\cite{devlin:bert}. In textual matching encoder part, we connect the two sentences end to end and send them to the pre-trained NLI model, and then get the text representation vector $r_t$. Then the visualization feature $r_{v,at}^{1}$ and $r_{v,at}^{2}$ is computed as Eq. (\ref{equ:att},\ref{equ:vta}) in section 3.1. We compute a hadamard product of visualization features and matching $p_{v}$ as 
\begin{equation}
p = \phi(r_{v,at}^{1} \odot r_{v,at}^{2}) + \phi(r_t)
\end{equation}
where $\phi$ is one hidden layer DNN with ReLU activations. 
The final output of this imaginative natural language inference model is category with maximum score. The objective function is the same as base model.

\subsubsection{Integration to Reading Comprehension}
In reading comprehension tasks, a passage usually contains multiple sentences which is denoted as $P={s_1, s_2, ... s_l}$.  Since the text representation model gives one representation for each passage, we apply a "late fusion" manner where the visual features of every sentences are combined together and then fed to the fusion network. As shown in Fig. \ref{fig:MRC}, in the integrated model, the three-way attention encoder generate a whole representation $r_p$ for $P$ as the base model~\cite{trian:2018}. Since each sentence in passage gets one image from the visualization mechanism, the features of these images are stacked together as $\hat{r_t} = [r_{1,v}, r_{2,v}, ... r_{n,v}]$. 
Passage representation $r_p$ then plays the role of computing attention weights of all stacked image features by Eq. (\ref{equ:att},\ref{equ:vta}) in section 3.1. The image representation $r_v$ is computed as in section 3.1. We denote the final score as $p$, and formulate the calculation as:
\begin{equation}
p = \phi(r_v) \cdot r_{qc} + p_t
\end{equation}
where $\cdot$ is dot product, $r_{qc}$ is the concatenation of question representation, $p_t$ is the logistic score of choice in the base model. The final output of this reading comprehension model is choice with maximum $p$ as in the base model ~\cite{trian:2018}.

\section{Experiments}

\subsection{Dataset}
\textbf{Natural Language Inference Dataset}
Stanford Natural Language Inference (SNLI) corpus~\cite{snli:2015} and SICK~\cite{SICK} are used in natural language inference evaluation. These two corpus contains human-annotated English sentence pairs on real world  pictures. The categories of both datasets are entailment, contradiction, and neutral. 

\textbf{Reading Comprehension Dataset}
SemEval-2018 task 11~\cite{semeval:2018} is used for reading comprehension evaluation, because this task requires commonsense knowledge about the world. It focuses on stories about simple and explicit in the description of script events and script participants. 

\textbf{Imagebase} The COCO dataset ~\cite{coco:2014} and Flicker 30k ~\cite{flicker} are used as imagebase. Since human visualization mechanism could recall the visual scene they used to see, both training set and develop set are used. There are 123K images in the COCO train and validation set. The Flicker dataset has 30K images. We use the smaller Flicker corpus as alternate small imagebase to evaluate the influence of imagebase size.

\subsection{Implementation}
During training Ours, we use Adam optimizer with 1e-4 learning rate and dropout rate of 0.4. When fine-tune the whole model the learning rate is set to 1e-5. In both step we train the model until accuracy stop increasing on develop set. The size of visual vector is the same as hidden size of the pre-trained NLU models to be integrated. 

\begin{table}
	\begin{center}
		\setlength\tabcolsep{3pt}
		\begin{tabular}{lcc} 
			\toprule
			\textbf{Models} & \textbf{Dev}& \textbf{Test} \\
			\midrule
			TriAN \cite{trian:2018}  & 82.78 & 81.08\\
			\midrule
			TriAN + ConceptNet \cite{conceptnet:2012} & 83.84 & 81.94\\
			TriAN + Ours& 83.90  & 82.10\\
			TriAN + Ours + ConceptNet& \textbf{84.42} & \textbf{83.12}\\
			\midrule
			TriAN + ConceptNet (ensemble) & 85.27 & 83.84\\
			TriAN + Ours  + ConceptNet (ensemble) & \textbf{85.94} & \textbf{84.41}\\
			\bottomrule
		\end{tabular}
	\end{center}
	\caption{Reading comprehension results on SemEval 2018 Task 11. }
	\vspace{-20pt}
	\label{tab:SemEval}
\end{table}

\subsection{Reading Comprehension}
The champion of the SemEval 2018 Task 11 TriAN \cite{trian:2018} is used as the reading comprehension baseline.
As the result in table.~\ref{tab:SemEval}, after integrating the proposed sentence visualization and fusion framework, we achieved better performance on both single and ensemble condition. We also compare the performance gain with ConceptNet~\cite{conceptnet:2012} which is reported in the TriAN \cite{trian:2018} paper. The sentence visualization and fusion framework brings 1.34\% improvement, higher than 0.82\% improvement of ConceptNet.

\begin{wraptable}{r}{0.4\linewidth}
	\vspace{-30pt}
	\begin{center}
		\setlength\tabcolsep{5pt}
		\begin{tabular}{lcc} 
			\toprule
			\textbf{Train Base}  & \textbf{Test Base} & \textbf{Test} \\
			\midrule
			\multirow{2}{*}{COCO}  & COCO  & 83.12\\
			& Flicker  & 83.09\\ 
			\midrule
			\multirow{2}{*}{Flicker}  & COCO  & 82.81\\
			& Flicker  & 82.81\\
			\bottomrule
		\end{tabular}
	\end{center}
	\caption{Result of SemEval 2018 Task 11 on 120k COCO  and 30k Flicker imagebases.}
	\vspace{-10pt}
	\label{tab:imgbase}
\end{wraptable}
\subsubsection{Impact of Imagebase Size} We test the performance of integrated model in different train/test imagebase. As shown in Table.~\ref{tab:imgbase}, using a bigger imagebase in train and test works better than a smaller imagebase. When the test image memory base shrinks to the smaller Flicker set, the performance reduces but is still higher than using the smaller image both in train and test. More importantly, when the model is trained on the Flicker30k imagebase, using a bigger one does not contribute to test performance. These results suggest that the framework works better when the size of the imagebase in train and test increases.

\subsubsection{Analysis by Question Categories}
To analyze how the sentence visualization and fusion framework helps reading comprehension, we performed an analysis of performance change of the TriAN o/w our framework on six most frequent question types. As the result in Table~\ref{tab:question}, the framework benefits most to \textit{what} and \textit{where} questions, and does not help answering abstract \textit{why}, \textit{who} and \textit{when} questions. According to our hypothesis, our mechanism tends to benefits the questions about the real world information, but could not help answering questions about reasoning and time.

\begin{table}
	\begin{center}
		\setlength\tabcolsep{3pt}
		\begin{tabular}{lcccccc} 
			\toprule
			\textbf{Models} & \textbf{y/n} & \textbf{what}& \textbf{why}& \textbf{who}& \textbf{where} &\textbf{when} \\
			\midrule
			TriAN        & 81.4 & 85.4  &  \textbf{84.9} & 89.7 & 84.7 & 78.9 \\
			TriAN + Ours   & \textbf{81.6} & \textbf{86.3}  &  84.4 & 89.7 & \textbf{85.3} & 78.9 \\
			\bottomrule
		\end{tabular}
	\end{center}
	\caption{Comparison of reading comprehension w/o our framework on six most frequent question types. }
	\label{tab:question}  
	\vspace{-30pt}
\end{table}

\begin{figure}
	\begin{tabular}{c c}
		\includegraphics[width=0.5\linewidth]{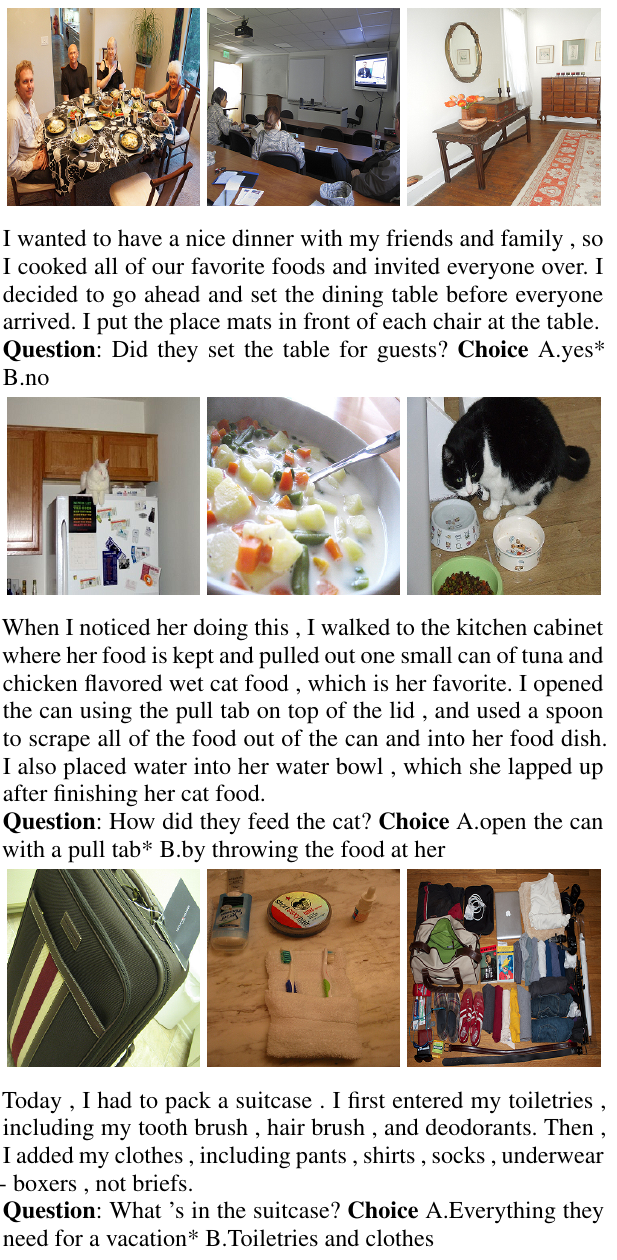} &
		\includegraphics[width=0.5\linewidth]{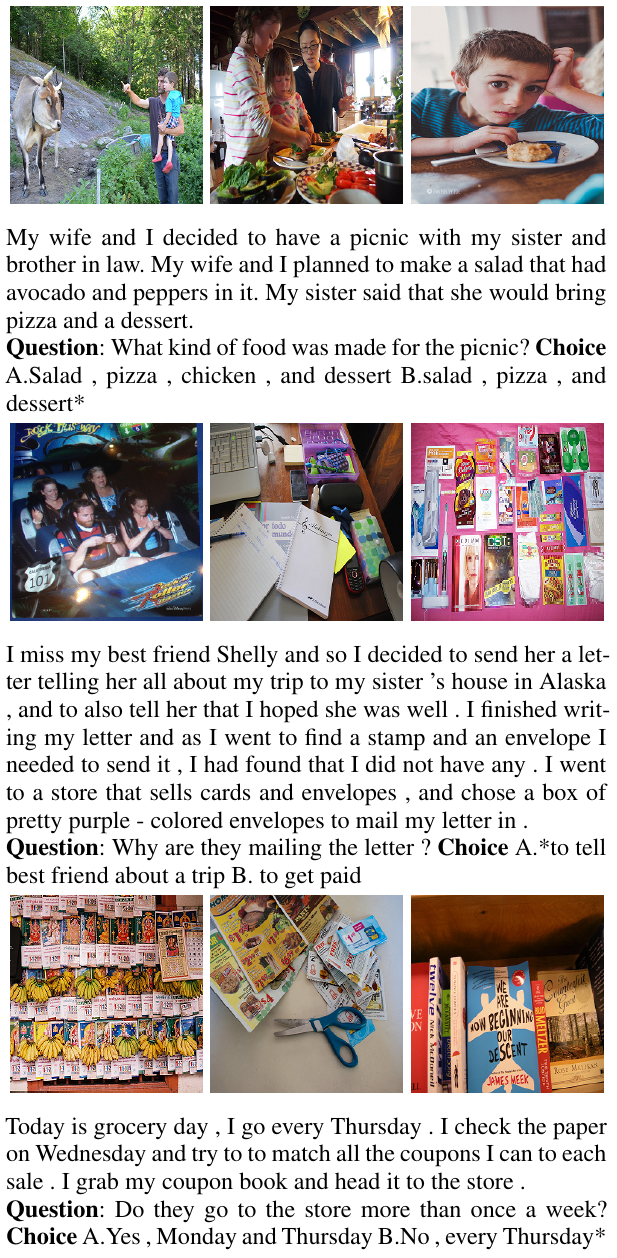} \\
		\small \textbf{Positive cases} & \small \textbf{Negative cases} \\
	\end{tabular}
	\caption{Cases our framework caused base model to switch from wrong to right and flipped the base model from right to wrong. The ground truth answer is marked with *.}
	\label{fig:case_pos}
	\vspace{-15pt}
\end{figure}
\subsubsection{Case Study}
To further analyze what the proposed framework provided to benefits reading comprehension task, we give three examples of integrating our framework causes the base model to switch from wrong to right in the Fig.~\ref{fig:case_pos}. In the cases, the helpful images contain environment objects (table, etc), characters (human, etc) and action (eat) information. It seems like the visualization tends to be helpful when direct clues (table, daily necessities) that match with the right answer appeared in the image. It can also be found that the framework does not provide proper images all the time or do not fully reflect the sentence content, which suggests that the framework may benefits from better image retrieval or generation models.

\begin{wraptable}{r}{0.5\textwidth}
	\vspace{-20pt}
	\centering
	\setlength\tabcolsep{2pt}
	\begin{tabular}{lcc}
		\toprule
		\textbf{Models} & \textbf{SNLI} (\%) & \textbf{SICK} (\%)\\
		\midrule
		*Cap2Both~\cite{2018:NAACL_sentence_visual} & - & 81.7 \\
		Picturebook \cite{hinton:2018}  & 86.5 & -\\
		MT-DNN~\cite{SNLI_1} & 91.1& -\\
		BERT \cite{devlin:bert} & 91.2 & 90.4\\
		\midrule
		BERT + Ours & \textbf{91.3} & \textbf{90.6}\\
		\bottomrule
	\end{tabular}
	\caption{Accuracy on SNLI and SICK. }
	\vspace{-20pt}
	\label{tab:nli}
\end{wraptable}
\subsection{Natural Language Inference}
Table ~\ref{tab:nli} shows the natural language inference results of integrating the mechanism to language inference model based on the pre-trained BERT~\cite{devlin:bert} model. Our framework improved the performance of the base model and achieved very high performance. Since our framework is designed specifically for pre-trained language models, the integrated model largely outperforms other multimodal models. It worth noting that the performance of integrated NLI model is largely depend the strong base model, so we then analyze how the proposed sentence visualization and fusion framework helps in the performance gain.

\subsubsection{Impact of retrieval score}
The performance change of SNLI samples with different retrieval scores are shown in the Table \ref{tab:nli_score}. The improvement increases following the retrieval score, which indicates that the proposed framework works better on concrete sentences that are more easily to be visualized. The result also suggests that framework is likely to work better as the text-to-image retrieval research develops in the future.
\begin{table}
	\vspace{-10pt}
	\centering
	\setlength\tabcolsep{10pt}
	\begin{tabular}{lcccc} 
		\toprule
		\textbf{Models} & \textbf{0.45+}  & \textbf{0.50+}  & \textbf{0.55+} & \textbf{0.60+} \\
		\midrule
		BERT    & 91.54\% & 92.07\% & 92.27\% & 90.08\%\\
		BERT + Ours & \textbf{91.63\%} & \textbf{92.22\%} & \textbf{92.44\%} & \textbf{91.14\%} \\
		\bottomrule
	\end{tabular}
	\vspace{5pt}
	\caption{Accuracy w/o our framework on SNLI samples with different retrieval scores.}
	\vspace{-30pt}
	\label{tab:nli_score}
\end{table}

\begin{figure*}
	\centering
	\setlength\tabcolsep{1.2pt} 
	\begin{tabular}{p{0.32\textwidth} p{0.32\textwidth} p{0.32\textwidth}} 
		\renewcommand\arraystretch{0.6}
		\subfigure[Entailment]{
			\begin{tabular}{p{0.31\textwidth}}
				\includegraphics[width=0.15\textwidth, height=1.8cm]{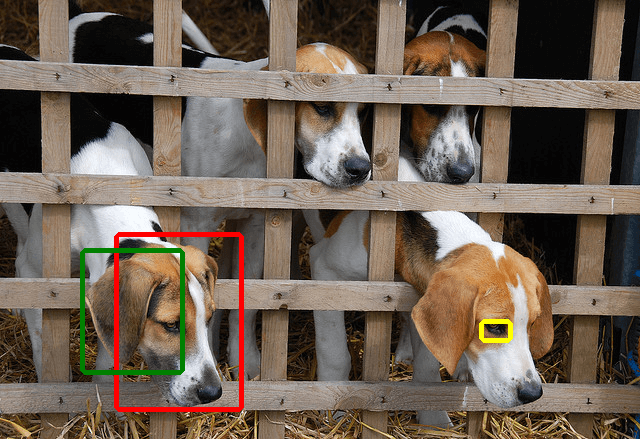}
				\includegraphics[width=0.15\textwidth, height=1.8cm]{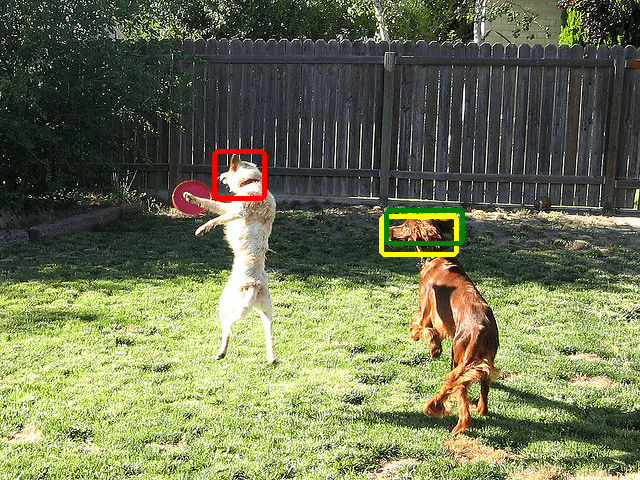} \\
				\setlength{\baselineskip}{7pt} {\scriptsize \textbf{S1:} Two dogs are looking through a rusty wire fence. } \\
				\setlength{\baselineskip}{7pt} {\scriptsize \textbf{S2:} Two dogs are in the yard. } \\ 
				{\scriptsize \textbf{BERT}: Neutral.} \\
				{\scriptsize \textbf{BERT + Ours: Entailment.}} \\
				
				\includegraphics[width=0.15\textwidth, height=1.8cm]{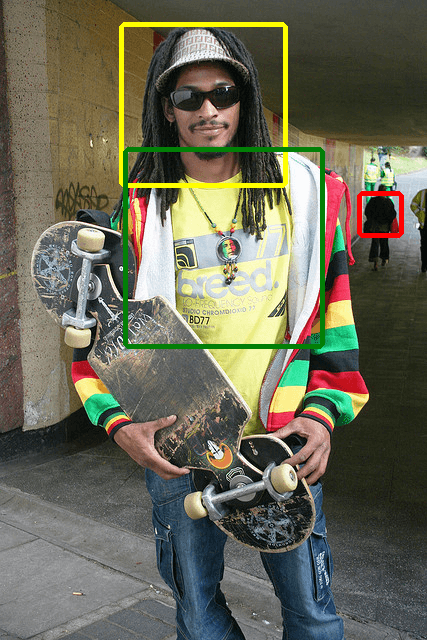}   
				\includegraphics[width=0.15\textwidth, height=1.8cm]{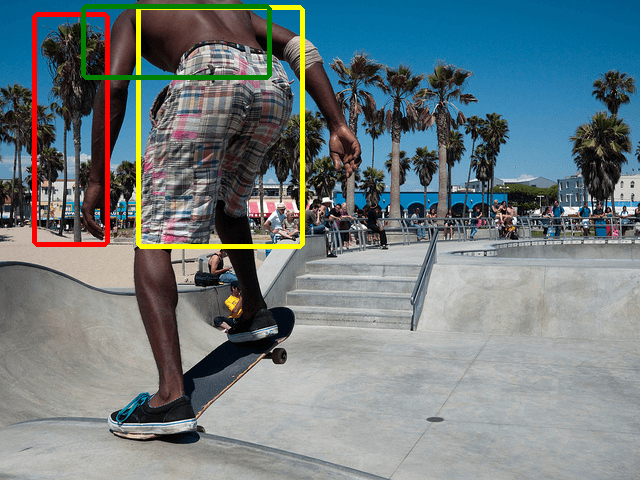} \\
				\setlength{\baselineskip}{7pt} {\scriptsize \textbf{S1:} A young man wearing black and carrying a skateboard looks at an outdoor skateboard park. } \\
				\setlength{\baselineskip}{7pt} {\scriptsize \textbf{S2:} A guy with a skateboard is looking at a park to skateboard. } \\ 
				{\scriptsize \textbf{\textbf{BERT}: Entailment.}} \\
				{\scriptsize \textbf{BERT + Ours}: Neutral.} \\
			\end{tabular}
		}
		&  
		\renewcommand\arraystretch{0.6}
		\subfigure[Neutral]{
			\begin{tabular}{p{0.31\textwidth}}
				\includegraphics[width=0.15\textwidth, height=1.8cm]{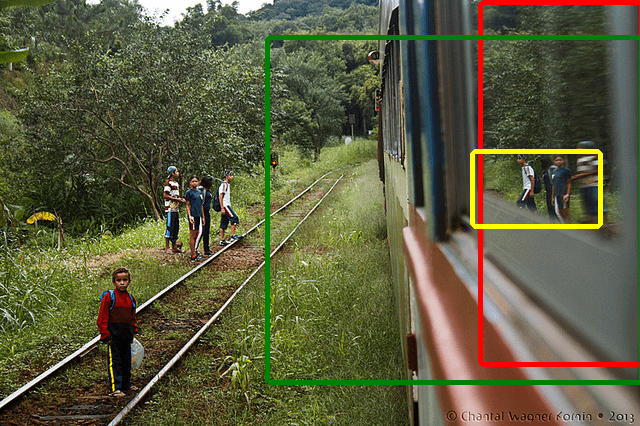}   
				\includegraphics[width=0.15\textwidth, height=1.8cm]{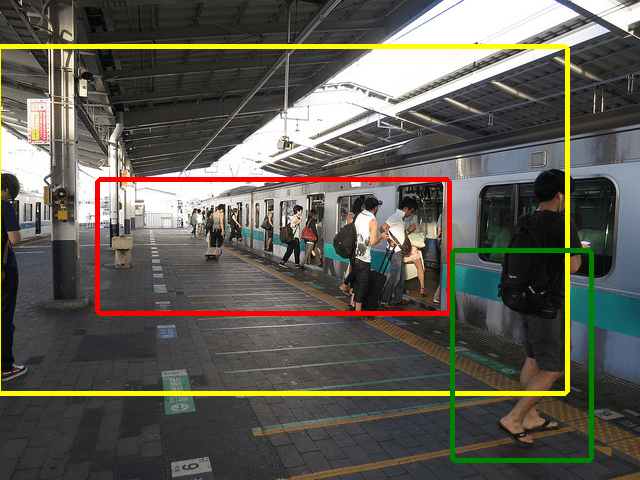} \\
				\setlength{\baselineskip}{7pt} {\scriptsize \textbf{S1:} People stand along a train track and take pictures. } \\
				\setlength{\baselineskip}{7pt} {\scriptsize \textbf{S2:} Some people are walking to the train station.} \\ 
				{\scriptsize \textbf{BERT}: Contradiction .} \\
				{\scriptsize\textbf{ Ours: Neutral.} } \\
				
				\includegraphics[width=0.15\textwidth, height=1.8cm]{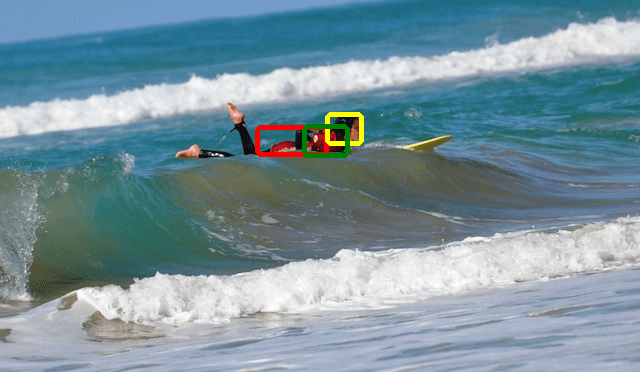}   
				\includegraphics[width=0.15\textwidth, height=1.8cm]{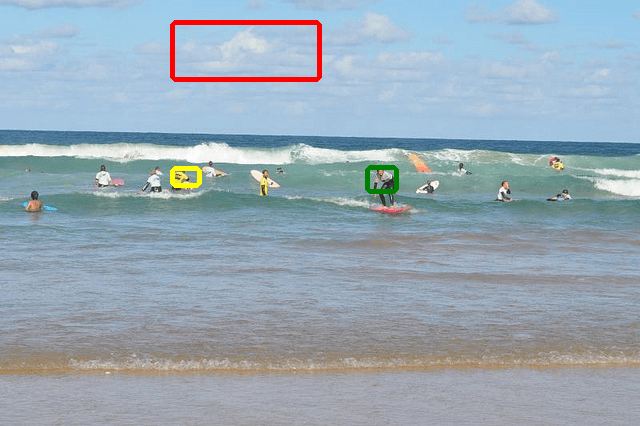} \\
				\setlength{\baselineskip}{7pt} {\scriptsize \textbf{S1:} Boys with their backs against an incoming wave. } \\
				\setlength{\baselineskip}{7pt} {\scriptsize \textbf{S2:} A group of people play in the ocean. \hfill \break } \\ 
				{\scriptsize \textbf{BERT}: Neutral.} \\
				{\scriptsize \textbf{BERT + Ours}: Entailment.} \\
			\end{tabular}
		}
		&  
		\renewcommand\arraystretch{0.6}
		\subfigure[Contradiction]{
			\begin{tabular}{p{0.31\textwidth}}
				\includegraphics[width=0.15\textwidth, height=1.8cm]{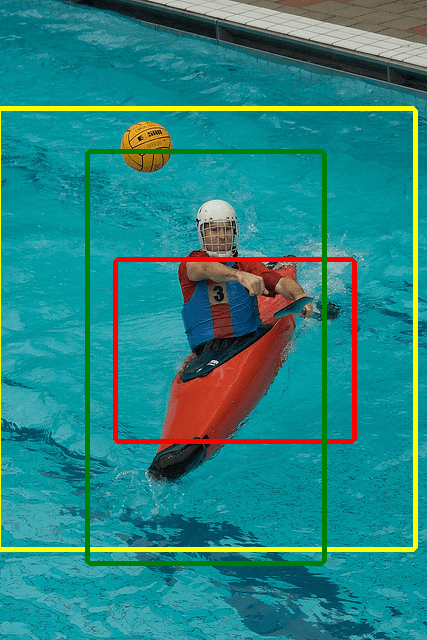} 
				\includegraphics[width=0.15\textwidth, height=1.8cm]{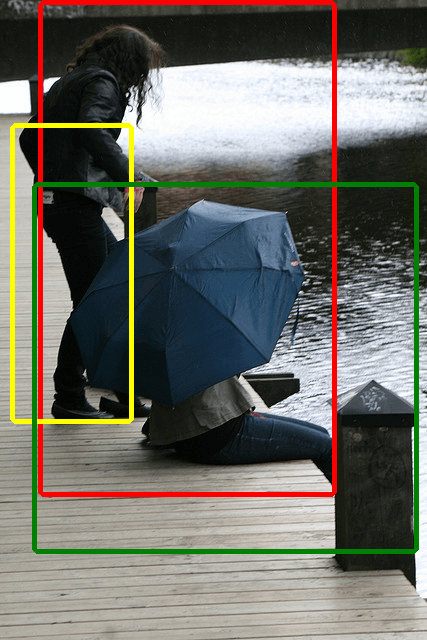} \\
				\setlength{\baselineskip}{7pt} {\scriptsize \textbf{S1:} A person swimming in a swimming pool.\hfill \break} \\
				\setlength{\baselineskip}{7pt} {\scriptsize \textbf{S1:} A person is trying not to get wet.} \\ 
				{\scriptsize \textbf{BERT}: Neutral.} \\
				{\scriptsize  \textbf{Ours:Contradiction}.} \\
				
				\includegraphics[width=0.15\textwidth, height=1.8cm]{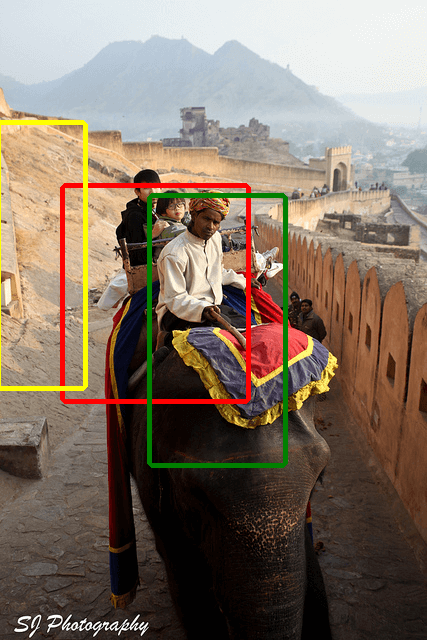}
				\includegraphics[width=0.15\textwidth, height=1.8cm]{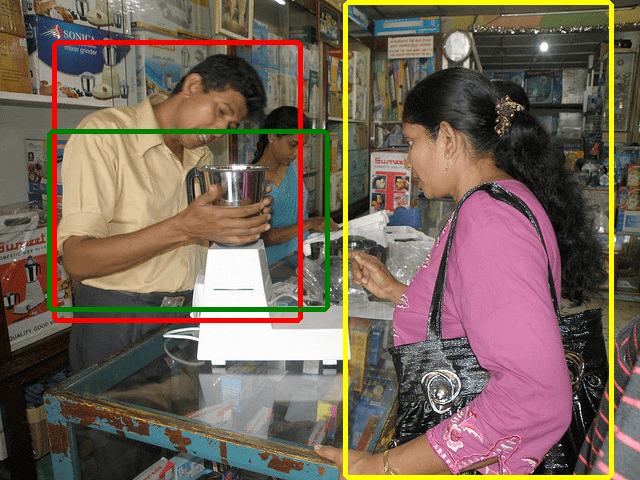} \\
				\setlength{\baselineskip}{7pt} {\scriptsize \textbf{S1:} A sungalsses vendor gazes into the distance. \hfill \break} \\
				\setlength{\baselineskip}{7pt} {\scriptsize \textbf{S1:} A vendor talks to a customer. \hfill \break}\\ 
				{\scriptsize \textbf{\textbf{BERT}: Contradiction}.} \\
				{\scriptsize \textbf{BERT + Ours}: Neutral.} \\
			\end{tabular}
		}\\
	\end{tabular}
	
	\caption{Sentence visualization with top-3 detected objects on SNLI. }
	\label{fig:case_snli}
	\vspace{-10pt}
\end{figure*}

\subsubsection{Cases Study}
Three good and bad examples for each categories are collected in Fig.~\ref{fig:case_snli}. In each case, the visualized image and top 3 object areas the model interests are labeled. In the image, the mechanism located entities (dog, people and human face) and tends to highlight the subject and object in the sentence. 
In good cases, the mechanism provides the state of objects, including position, action, and appearance, which is not given in the sentence in detail. In the bad cases, the images do not exactly match the sentence, especially when there are many quantifier or adverbials in the sentence, likely because these words impose too many restrictions that beyond the limitation of imagebase.


\section{Conclusion}
In this paper we proposed a sentence visualization and fusion framework and used it to improve pre-trained language inference and reading comprehension models. Unique from existing multimodal representation learning models, we focused on providing a flexible text-visual fusing approach that can be easily integrated into pre-trained language models. The proposed approach brings improvements to strong baseline models on the public corpora SNLI, SICK and SemEval 2018 Task 11. Further analysis and case studies also show the reasonableness of performance gain acquired by the visualization.

Several future directions and improvements could be considered. Our main objective was to show the potential of sentence visualization. We found that the proposed framework works well when images are correctly retrieved; future work might consider using stronger image retrieval networks. The limitation of imagebase size may be released through generation models.
\bibliographystyle{splncs04}
\bibliography{acl2018}

\end{document}